# Planning with External Events

**Jim Blythe**
School of Computer Science
Carnegie Mellon University
Pittsburgh, PA 15213
blythe@cs.cmu.edu

## Abstract

I describe a planning methodology for domains with uncertainty in the form of external events that are not completely predictable. The events are represented by enabling conditions and probabilities of occurrence. The planner is goal-directed and backward chaining, but the subgoals are suggested by analysing the probability of success of the partial plan rather than being simply the open conditions of the operators in the plan. The partial plan is represented as a Bayesian belief net to compute its probability of success. Since calculating the probability of success of a plan can be very expensive I introduce two other techniques for computing it, one that uses Monte Carlo simulation to estimate it and one based on a Markov chain representation that uses knowledge about the dependencies between the predicates describing the domain.

## 1 INTRODUCTION

One of the central assumptions of classical AI-based planning is that the state resulting at some time after performing an action can be predicted completely and with certainty. This assumption permits a style of planning in which a goal, represented by a sentence in first-order logic, is achieved exactly by a plan, represented as a partially ordered set of actions. The plan need include no sensing or branching because of the assumption.

More realistic planners allow for uncertainty in their domains of application. Specifically, uncertainty in the domain is typically represented in one or more of 3 ways:

1. non-deterministic effects of operators, possibly with probability distributions,
2. uncertainty about the initial conditions of the problem, and
3. uncertainty about future states due to unpredictable external events.

This classification is not intended to be exhaustive or exclusive. For instance, non-deterministic effects of operators in the Cassandra planner are represented by conditional effects dependent on unknown (and unknowable) initial conditions [Pryor & Collins, 1993].

In a domain that includes uncertainty, plans must account for the different possible outcomes of actions. Thus plans take on the form of a Markov chain, rather than a partially ordered set of actions [Dean *et al.*, 1993, Thiebaux *et al.*, 1993]. Each node of the chain represents a set of states and the action that the plan associates with this set. The arcs, with associated probabilities, represent the different possible outcomes of the action. Plans may also need to include explicit sensing, to determine which alternative was reached.

Plans produced in uncertain domains may not achieve their goals with certainty. Planners dealing with uncertainty have typically considered the probability of plan success as a measure of goodness of the plan [Kushmerick, Hanks, & Weld, 1993], as is done here. When there are no time constraints on the planning process, one might continue planning until some threshold probability of success for the plan is reached [Kushmerick, Hanks, & Weld, 1993], or attempt to cover every possible outcome [Pryor & Collins, 1993, Mansell, 1993]. For cases when the time available for planning might preclude this, "anytime algorithms" that increase the expected utility of the plan over time have been proposed [Dean *et al.*, 1993, Koenig, 1992].

Most planners that deal with uncertainty only consider uncertain action effects or uncertainty in the initial state. In this paper I describe planning under uncertainty due to external events which are not completely predictable. The planner uses an explicit, probabilistic representation of external events to reason about the way its domain can change. Events are represented similarly to operators, with enabling conditions and a single, fixed probability of occurrence over a discrete time unit[1]. This representation allows the use of techniques similar to classical planning to reason about events and design plans that have a high probability of success in eventful domains. It also seems intuitively plausible that some knowledge of the way a domain is likely to change would be available to a planner that could be encoded probabilistically. Other planning formalisms are able to represent events more generally [Dean & Kanazawa, 1989, Haddawy, 1991], but they do not specify planning mechanisms that make use of the information directly.

[1]Continuous time representations have also been considered, but this aspect does not affect the planner, since it is not yet able to handle actions whose effects depend on metric time

| | |
|---|---|
| Event: | drive |
| Duration: | 1 |
| Types: | ?taxi: Taxi |
| | ?source, ?dest: Place |
| Preconditions: | (and (same-city ?source ?dest) |
| | (location ?taxi ?source)) |
| Initial Deletes: | ((location ?taxi ?source)) |
| Initial Adds: | ((location ?taxi on-the-road)) |
| Final Deletes: | ((location ?taxi on-the-road)) |
| Final Adds: | ((location ?taxi ?dest)) |

Figure 1: The operator schema drive

| | |
|---|---|
| Event: | lose-package-from-airport |
| Duration: | 0 |
| Types: | ?package: Package |
| | ?airport: Airport |
| Preconditions: | (and (not (protected ?package)) |
| | (location ?package ?airport)) |
| Deletes: | ((location ?package ?airport)) |
| Adds: | ((location ?package lost)) |
| Probability: | 0.1 |

Figure 2: The event schema lose-package-from-airport

I describe a method of plan repair for dealing with this type of uncertainty that uses goal-directed search, where the goals are suggested by decision-theoretic criteria. I argue in section 6 that this is a promising way to combine the approaches of decision theory and AI-based plan synthesis. In the next section I describe the representation for external events in more detail, and discuss the planning algorithm, which is implemented on top of Prodigy 4.0 [Carbonell & the PRODIGY Research Group, 1992] and IDEAL [Srinivas & Breese, 1990]. In section 3 I illustrate the technique on an example domain. In the next section I describe how the Bayesian belief net is constructed from the plan to reason about events and in section 5 I discuss the separate methods developed for estimating the probability of success of a plan or computing it more efficiently.

# 2 THE PLANNING REPRESENTATION AND ALGORITHM

## 2.1 REPRESENTATION OF ACTIONS AND EXTERNAL EVENTS

In this paper, I ignore uncertainty due to incomplete information about the planner's current state, or due to nondeterministic effects of operators, and concentrate exclusively on uncertainty due to external events. The planner derives its state and operator representation from Prodigy 4.0. States are described in a typed predicate logic. Operators are represented by preconditions and use add and delete lists to describe their effects. Preconditions are statements of first-order logic, and effects can include conditions, and be universally quantified. In addition, each operator has a fixed duration, which is an integer.

Operators with non-zero duration use two sets of add and delete lists in order to capture the state of the world while the operator is being applied. For example, the operator drive shown in figure 1 has the global effect of moving a taxi ?taxi from a source location to a destination. The symbols ?taxi, ?source and ?dest are typed variables. It is modelled as first moving the taxi from the source location to the location "on the road", and on completion of the operator one time unit later, moving it from "on the road" to the destination. The planner ignores the intermediate state and treats the operator as an atomic action, while routines that reason about external events consider the intermediate state. In this way the system can reason about operators that take time to apply, but does not introduce the full complexity of continuous change.

Events are represented in the same way as operators, but storing in addition the conditional probability that the event will take place given that the preconditions are satisfied. For example, figure 2 describes the event that a package is lost from an airport. This event can be read as "any package can be lost with probability 0.1 during any time unit in which it is left unprotected at an airport".

Events may only take place during execution of operators of non-zero duration. If an event and an action complete at the same time, the effects of the event take place first, and then those of the action. One consequence of this is that if the event and action set the same state variable to a different value, the action's value will prevail.

A more general representation, that assigns a probability distribution to each event over the set of all states, might be preferable. The simplification used here, in which the event has a fixed probability in all the states satisfying the preconditions, and zero probability elsewhere, simplifies the use of goal-directed reasoning and backward chaining as discussed in the next section. More general models of external events have been proposed, e.g. [McDermott, 1982, Haddawy, 1991, Kanazawa, 1992], but these models have not been provided with planning mechanisms.

## 2.2 PLANNING ALGORITHM

The planner, as currently implemented, operates in a loop as follows:

Generate an initial plan $P$, to achieve a state satisfying the goal description.
Create a Bayesian net $B$ to represent the plan.

Loop:

If $B$ satisfies the threshold goal probability, return $P$.
Otherwise
  Augment $B$ to represent external events, and find a set of potential failures $F$. Each "failure" is an event node in $B$ and a related chain of events that makes the failure possible.

  Choose a failure to work on.

  Choose one of the following 3 repair methods:
    Create a plan to achieve a goal state from the state produced by the event and add this to $P$ and $B$ as a conditional branch, should the event sequence take place.

    Pick an event in the chain of events, and attempt to protect $P$ by adding steps to negate the preconditions of this event at the time it may take place

    Reduce the probability of the chain of events by moving the steps of the plan along a time line, and perhaps re-ordering them if consistent with goal satisfaction, in order to reduce the time over which the events can occur.

  Go to Loop.

Essentially this approach implements an abstraction barrier between the planner and the routines that reason about plan execution to return the set of potential failures $F$. The planner ignores external events, except those which are incrementally brought to its attention by the routines that augment the Bayesian net. These routines in turn only worry about those events that may affect the plan given to them.

Prodigy 4.0 is used to produce initial plans, ignoring all external events. In order the find a suitable plan, the planner may have to backtrack over all possible ways to fix each failure, as well as try different initial plans. Thus it finds a plan that will work if no failure events take place.

An interesting feature of this algorithm is the step that rectifies a failure by negating the preconditions of an event. This allows the planner to make use of its explicit knowledge of the causes of external events to formulate subgoals and create plans to defeat them. An example is the subplan that is generated in the next section to prevent the package from getting lost at the airport.

Calculating the complete set of event sequences that may defeat the plan is NP-hard, by a reduction from classical planning. The implemented system calculates a subset of the event sequences that may defeat a plan, as discussed in section 4.

# 3 AN ILLUSTRATION

## 3.1 EXAMPLE DOMAIN

Consider the following simple transportation problem. A number of packages are to be moved between locations, using combinations of taxis and airplanes. The taxis are used to move objects around within a city, and airplanes to move objects between airports of different cities. The most important predicates of the domain, with the types they expect, are location(object, location), lost(object), protected(object) and have-key(locker). Object is a super-type of package and vehicle, which is a super-type of taxi and airplane. Vehicle is also a sub-type of location.

There are nine operators:

```
load-taxi(?package, ?taxi, ?loc)
  Preconditions: (and (location ?package ?loc)
                      (location ?taxi ?loc))
  Adds:          ((location ?package ?taxi))
  Deletes:       ((location ?package ?loc))

unload-taxi(package, ?taxi, ?loc)
  Preconditions: (and (location ?taxi ?loc)
                      (location ?package ?taxi))
  Adds:          ((location ?package ?loc))
  Deletes:       ((location ?package ?taxi))

open-locker(?locker, ?funds)
  Preconditions: (have-money ?funds)
  Adds:          ((have-key ?locker)
                 (have-money (- ?funds 1)))
  Deletes:       ((have-money ?funds))

store(?package, ?locker, ?airport)
  Preconditions: (and (location ?package ?airport)
                      (location ?locker ?airport)
                      (have-key ?locker))
  Adds:          ((location ?package ?locker)
                 (protected ?package))
  Deletes:       ((location ?package ?airport))

unstore(?locker, ?airport)
  Preconditions: (and (location ?locker ?airport)
                      (have-key ?locker))
  Adds:          ((location ?package ?airport)
                 ∀ ?package with (location ?package ?locker))
  Deletes:       ((location ?package ?locker) ∀ ?package
                 (have-key ?locker))
```

In addition to drive shown in figure 1 there are 3 operators, load-airplane, unload-airplane and fly, which are analogous to load-taxi, unload-taxi and drive respectively, except that airplanes are constrained to move between airports while taxis are constrained to stay in the same city. Drive and fly are the only operators with non-zero duration, drive taking 1 hour and fly taking 5 hours. Since the current model of a plan does not allow simultaneous effects, the temporal length of a plan is uniquely determined by these steps.

The three events modelled in the domain are lose-package-from-airport, shown in figure 2, lose-package-from-post-office, which is analagous but has the lower probability of 0.05, and taxi-moves, which models a taxi moving between the locations in a city — each taxi will move to the other location with probability 0.2.

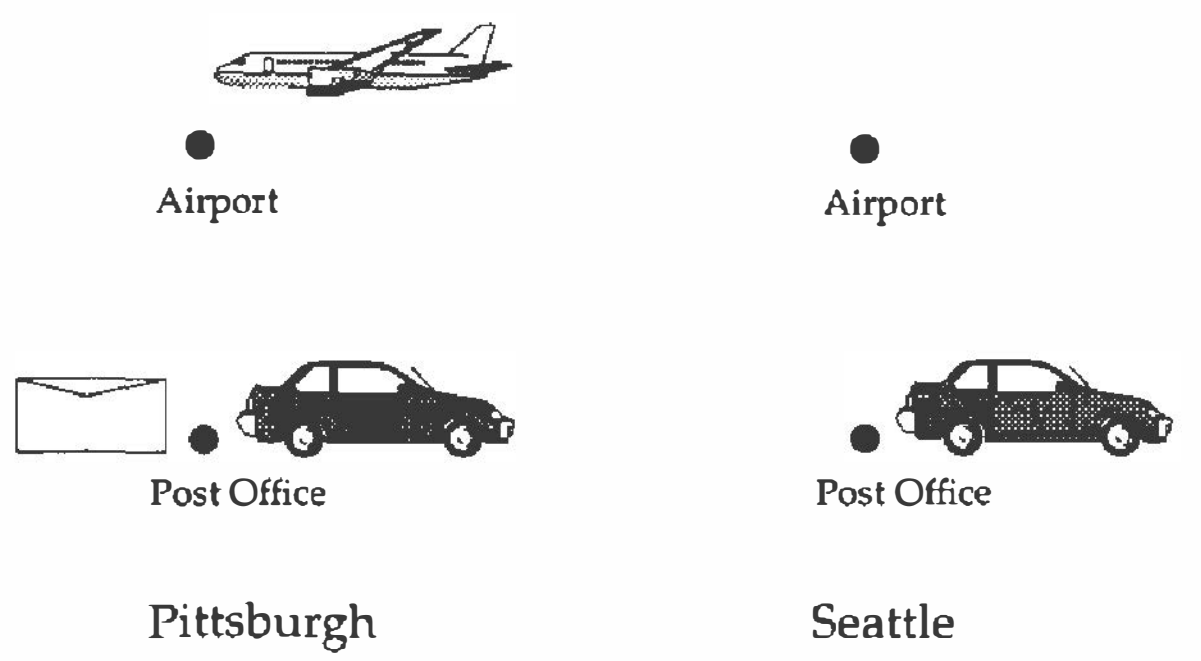

Figure 3: Initial state for the example problem

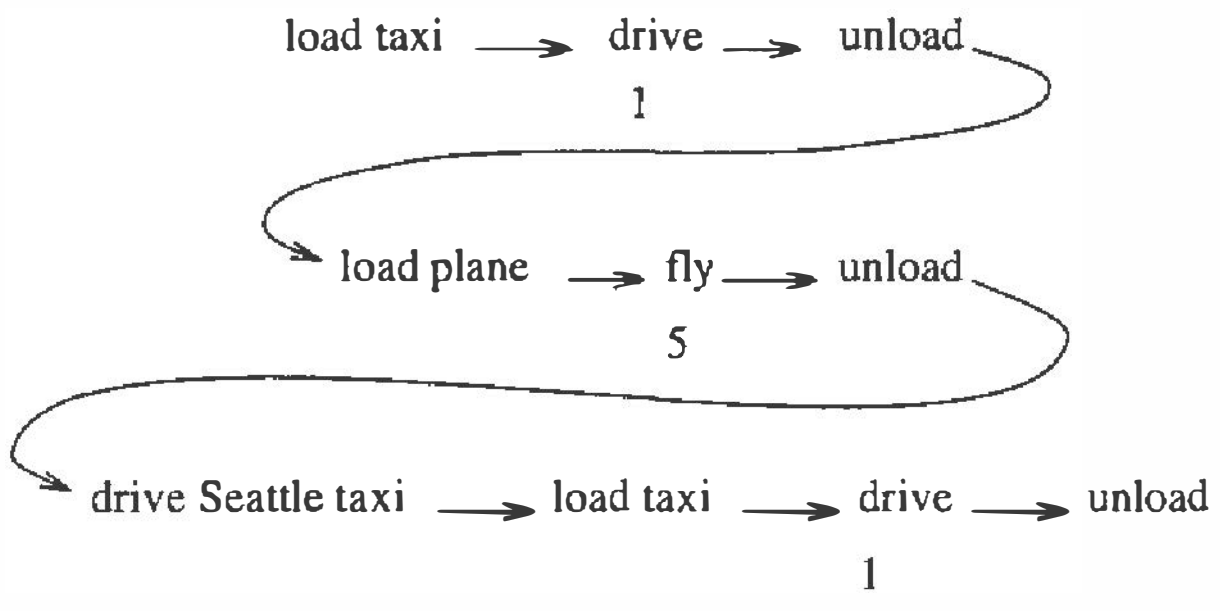

Figure 4: Initial plan generate by Prodigy

### 3.2 EXAMPLE PROBLEM

The initial state for the example problem is shown in figure 3. There is a package at the post office in Pittsburgh, where there is also a taxi. There is an airplane at the Pittsburgh airport and a taxi at the post office in Seattle. The goal is to have the package in the post office in Seattle.

Prodigy generates the initial plan for this problem that is shown in figure 4. A belief net is built from the plan which is analysed for possible failures. The routine used looks for event sequences of increasing length, first returning all one-event sequences that can defeat the plan. In this case, two are found. Firstly, the step to drive the taxi in Seattle from the post office to the airport can fail, because the taxi moved to the airport at some previous time. Secondly, the package could be lost from the Seattle airport while the Seattle taxi is in transit.

Next, probabilities are assigned to the events indicated in each failure by evaluating the belief net. These probabilities reflect all possible event sequences that can lead to the failure events. For example, in the first failure, we have the probability of an event sequence in which the taxi in Seattle has moved an odd number of times by the time the plan requires it to be driven to the airport. Methods for computing or estimating this are discussed in the next section. In this example, the probability that the taxi is in the wrong place is computed as 0.477, and the probability that the package is lost is 0.1.

The planner works on the highest probability failure first. Of its 3 possible methods, only re-planning from the undesirable outcome will solve this problem, and the planner adds a branch to the point in the plan before moving the taxi in Seattle, effectively making the action optional. The failures are now recalculated yielding only the second failure, the package being lost at the airport, with the same probability as earlier. No action can be taken if the package is lost, so the system tries to prevent the event from taking place. This is done by adding preconditions to the actions during which the event can take place (in this case driving the taxi) that negate the event's preconditions and re-planning. This method, which is similar to a standard technique for dealing with conditional effects [Pednault, 1988], leads to the final plan shown in figure 5.

## 4 BELIEF NET CONSTRUCTION

In order to perform the probability calculations efficiently, a Bayes net is automatically constructed from the plan. This representation has a number of advantages for reasoning probabilistically about plans. It makes efficient use of the dependency structure between domain features when computing probabilities, and has a sound theoretical basis [Pearl, 1988]. It can also efficiently maintain beliefs about unobservable world features based on observations during plan execution. The Bayes net is constructed in two stages. First, nodes are created representing beliefs in the features of the state that are relevant to the plan regardless of any external events that may take place. In the second stage nodes are added that represent external events and the state features that are relevant to them. The algorithm in detail is as follows:

$T = 0, S = 0$
*Stage 1:*
For each action $A$ in the plan
  Create a node $N_A$ to represent $A$ at time $T$ and stage $S$.
  For each precondition $P$ of $A$
    Find or create a node $N_P$ for $P$ at time $T$ and stage $S$.
    Link $N_P$ to $N_A$.
  If the $d$, the duration of $A$, is not 0 then
    $T = T + d, S = 0$
  otherwise $S = S + 1$
  For each effect $E$ of $A$
    Find or create a node $N_P$ for $P$ at time $T$ and stage $S$.
    Link $N_P$ to $N_A$.
*Stage 2:*
For each node $N$ in the belief net representing a state feature
  If $N$ is not the effect of an action,
    link it to the most recent previous node of the same type.
    Find the set $\mathcal{E}_\mathcal{N}$ of events that can affect $N$
    For each event $\epsilon$ in $\mathcal{E}_\mathcal{N}$
      Add nodes for the event and its preconditions
      in the same way as stage 1.
If new events were added, go back to *Stage 2*.

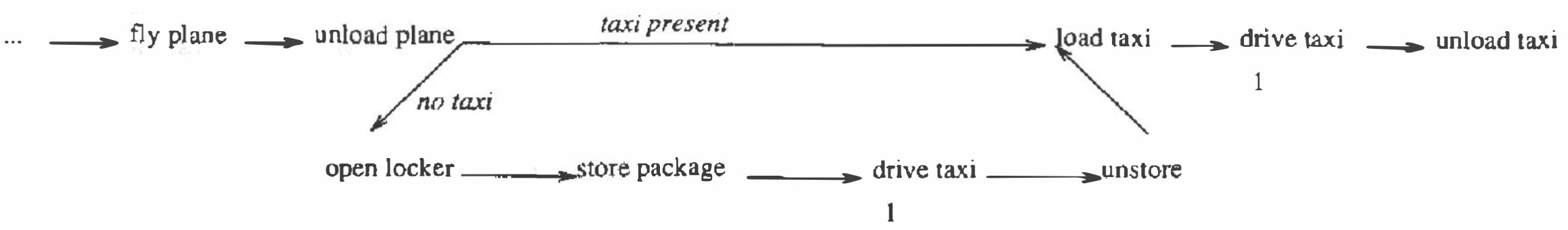

Figure 5: Final portion of the final plan

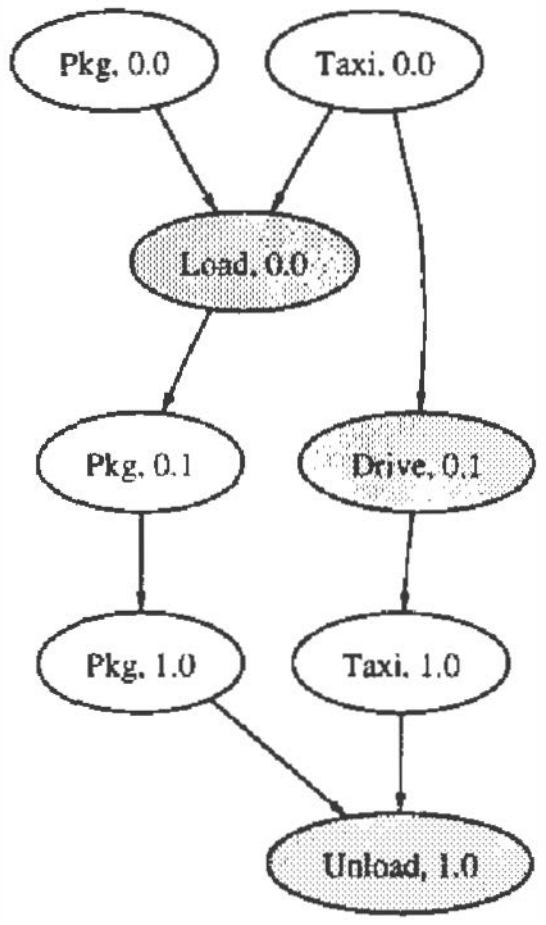

Figure 6: A portion of the Bayesian net constructed to reason about the initial plan, before events are considered. Shaded nodes represent actions. The order of action execution is downwards through the net.

This method leads to a belief net which is concise in the sense that only state features and events known to be relevant to the plan are represented. Since the only uncertainty considered in this paper stems from external events, the "net" after the first stage is completely deterministic.

## 4.1 STAGE 1: CONSTRUCTING A NET FROM THE PLAN

A portion of the Bayesian net that is constructed from the initial plan of the example is shown in figure 6, consisting of all nodes created up to and including the third action in the plan. Clear nodes represent the values of predicates at a particular stage in the plan, as represented by the numbers in the node label. The integer before the point is the time that the node refers to, and the integer after the node is the stage, which is incremented after each operator application. Thus the node labelled "Pkg, 0.1" represents the location of package1 after the first action taken at time 0 in the plan's execution. Shaded nodes represent the execution of actions in the plan.

Bayesian nets can be constructed from plans in a variety of ways, reflecting different styles of plan execution and monitoring, as well as the different forms of uncertainty in the domain representation [Blythe, 1994]. In this construction I assume that actions are executed in the linear sequence of the plan if possible, and if an action cannot be executed, the plan is considered to have failed. To illustrate how this is reflected in the net, consider the part of the net related to the first step in the plan, (load package1 pgh-taxi), represented by the node labelled "Load, 0.0". The location of the package after the load action is attempted is considered to be independent of its location before the action given that we know if the action was successful. This is because if the action is successful its new location is determined by the action alone, and if it is unsuccessful, the new location is set to a null value, signalling that the plan has failed. Since the plan is justified [Knoblock, 1991], the null value will propagate through to the node representing the goal state.

## 4.2 STAGE 2: REPRESENTING EXTERNAL EVENTS

Once the net representing the plan is completed, it is searched for points where external events might affect it. By the definition of events as discussed in section 2.1, these are limited to feature nodes that do not have an action as a parent. These nodes are the points in the plan where persistence is assumed. In figure 6, for example, the link between the nodes labelled "Pkg, 0.1" and "Pkg, 1.0" represents the system's belief that the package remains in the taxi while it is driven to the airport. Conceivably an event might interfere with this persistence link, although no such events are known to the system.

The algorithm therefore proceeds by searching from each node representing a persistence assumption for events whose effects would alter the value of the node and whose preconditions might be matched over the interval of the persistence. On finding such events, new nodes are added for the event and its preconditions. The nodes representing the events are not deterministic, reflecting the probability of occurrence assigned to the event. The nodes representing their effects are deterministic, representing persistence unless the event takes place. To find all sequences of events that might destroy a persistence assumption, this algorithm must be applied recursively on the enlarged net until no new events are discovered. The belief net that results from this is shown in figure 7.

Standard algorithms may be used to compute the probability of plan success from the belief net. However this is not tractable for reasonably sized problems, both because of the exhaustive search required to produce the full Bayesian net and because evaluating a Bayesian net is NP-hard in general [Cooper, 1990]. For this reason the algorithm presented here for producing the net represents an ideal case, while in general a number of short-cuts are used. These are discussed in the next section. In addition, work is in progress to find ways of producing the net along with special-case

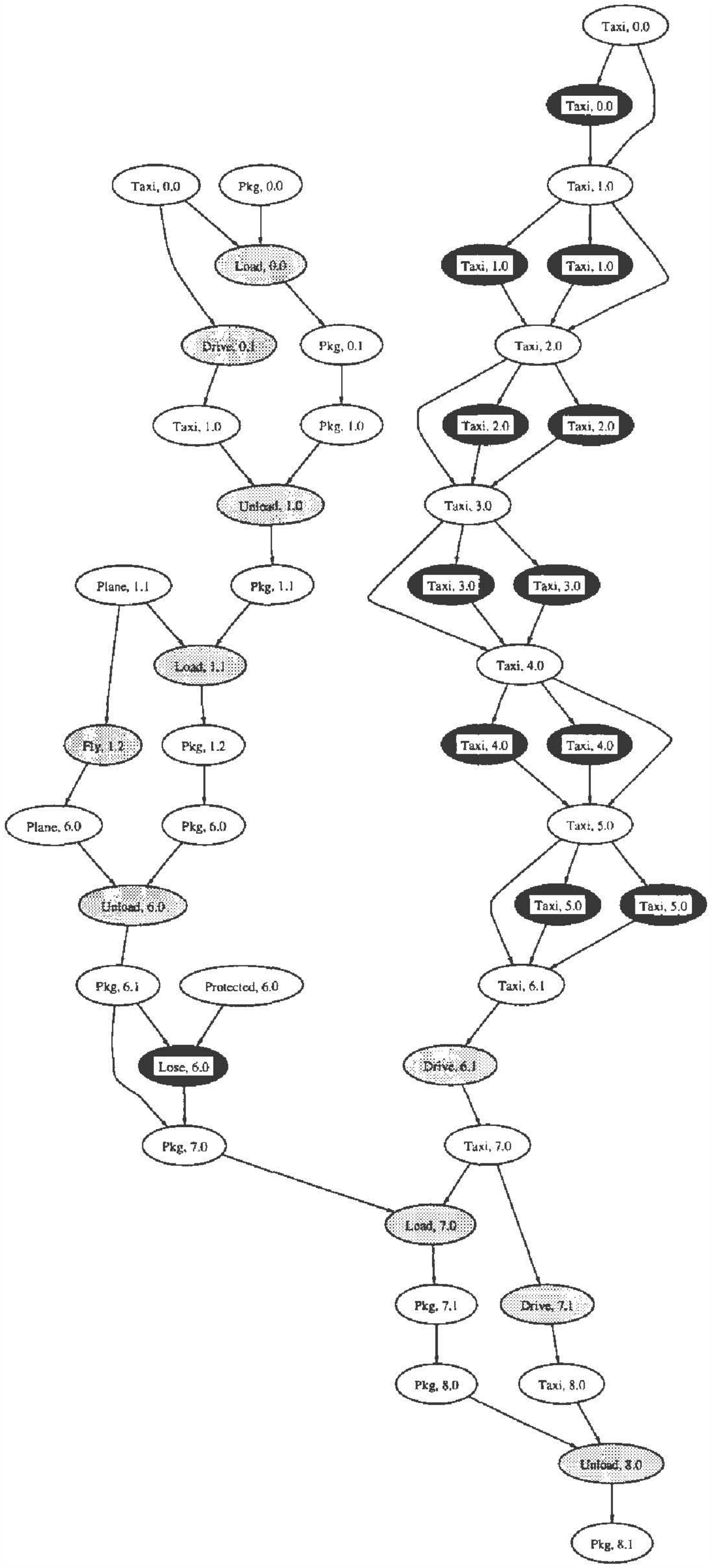

Figure 7: The Bayesian net for the initial plan with all relevant events. Shaded nodes represent actions in the plan and black nodes represent events that may affect it.

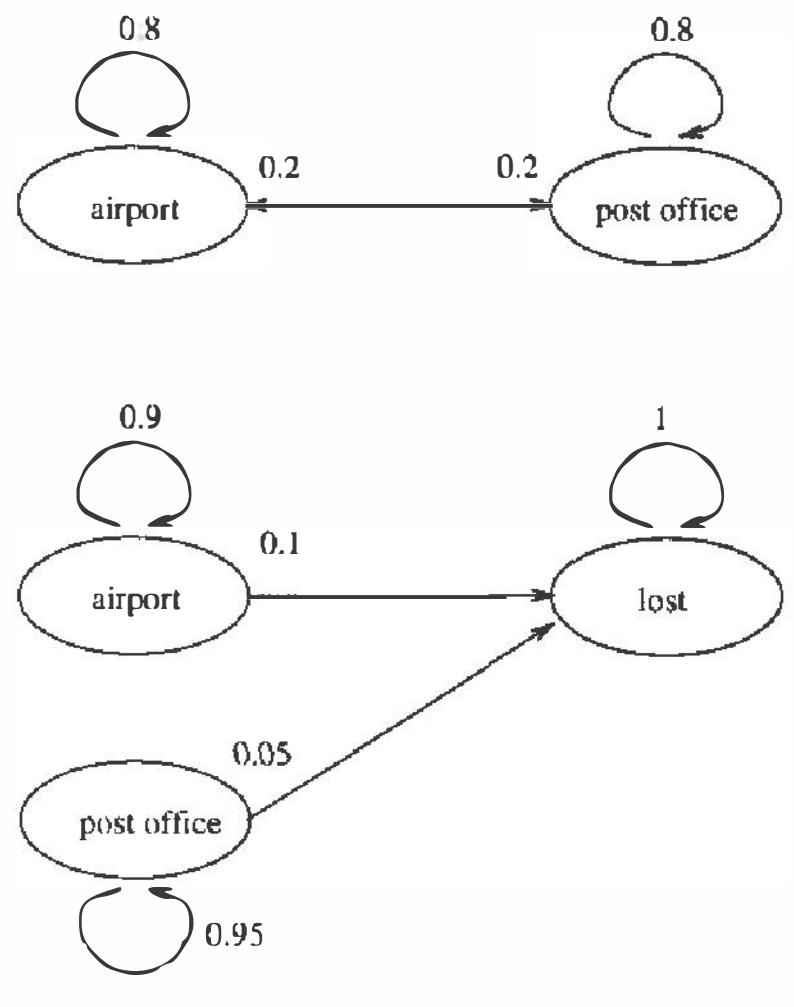

Figure 8: Markov chains for taxis and for packages

computational techniques that will improve efficiency.

## 5 ESTIMATING THE PROBABILITIES OF FAILURE SEQUENCES

Two methods have been investigated to reduce the time building and evaluating the belief net for the plan. The first is based on knowledge about independence of domain features. It removes parts of the net without altering the beliefs for the remaining nodes. The second by-passes the belief net by performing a Monte Carlo analysis. It is more broadly applicable than the first technique but is slow to converge.

### 5.1 EXPLOITING INDEPENDENCE

The important probabilities in the example problem can in fact be calculated very quickly if information about the independence of the different event types is supplied to the system. In this example each package is lost independently of the other packages and independently of the whereabouts of taxis, and each taxi moves independently of the other. We use this information and the Markov assumption implicit in the event specification to factor the domain into a number of small, independent Markov chains that completely describe it. These are shown in figure 8.

The initial plan and its two potential failure types require two probabilities to be calculated: the probability that the taxi in Seattle moves from the post office to the airport over 6 time units, and the probability that the package is lost from the airport in 1 time unit. Since in this case the Markov chains all have one or two states, these values can be computed analytically, and are approximately 0.477 and 0.1 respectively. If the Markov chains had more states, the Kolmogorov equations could be used for an efficient solution [Ross, 1980].

The Markov chain representing the taxi in Seattle can be clearly seen from the belief net in figure 7. This indicates

Checking initial plan:
checking step (load-taxi package1 pgh-taxi pgh-po)
deleting (location package1 pgh-po)
adding (location package1 pgh-taxi)
checking step (drive pgh-taxi pgh-po pgh-airport)
deleting (location pgh-taxi pgh-po)
adding (location pgh-taxi on-the-road)
step begun
deleting (location pgh-taxi on-the-road)
adding (location pgh-taxi pgh-airport)
step completed.
*Two uneventful steps are skipped here*
checking step (fly airplane1 pgh-airport seattle-airport)
deleting (location airplane1 pgh-airport)
adding (location airplane1 in-flight)
step begun.
** event taxi-moves takes place at tick 5.
deleting (location seattle-taxi seattle-po)
adding (location seattle-taxi seattle-airport)
deleting (location airplane1 in-flight)
adding (location airplane1 seattle-airport)
step completed.
*unload-airplane completes successfully*
checking step (drive seattle-taxi seattle-po seattle-airport)
precondition (location seattle-taxi seattle-po) is false
*** step was not applicable

Figure 9: A trace of plan simulation

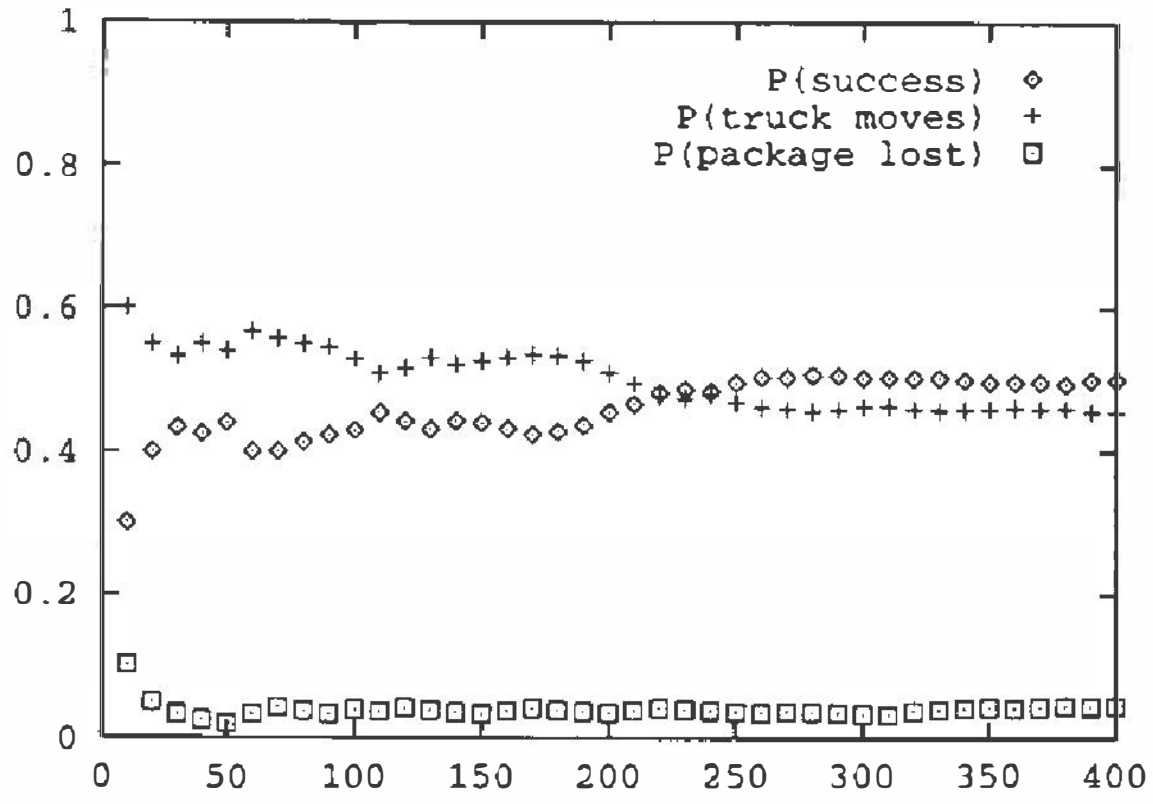

Figure 10: Convergence of the Monte Carlo simulation for the example problem

that the information about independence required for this technique could be learned analytically or empirically from analysis of belief nets constructed without this information.

### 5.2 MONTE CARLO SIMULATION

An alternative technique to building the belief net is a Monte Carlo simulation. The plan is simulated a number of times, each time allowing events to take place randomly according to their given probabilities. If an action in the plan has duration $n$, then all $n$ time steps are simulated. The model makes use of the intermediate states of operators as discussed in section 2.1. Essentially, the Monte Carlo technique is used to approximate the construction of the net and its simultaneous evaluation.

Figure 9 shows a trace run of the simulator with the initial plan from the example in the previous section. We see that execution proceeds normally until the last moment of the plane's flight to Seattle, when the taxi moves to the airport in Seattle and the plan fails (it is initially brittle). Each time a failure such as this occurs, the failure type consisting of the step that failed, each precondition that failed and the event that caused it is noted. Ultimately the proportion of times each type is implicated in failure will converge to the probability that this failure sequence will occur, and the proportion of times the execution succeeds will converge to the probability of plan success. By the central limit theorem, these estimates as random variables will converge to a normal distribution, with the desired mean and variance proportional to $1/\sqrt{N}$, where $N$ is the number of trials. Figure 10 graphs the convergence of our estimates for the probability of success and the two failure types for the example problem. It can be seen that this technique can very quickly tell us which is the more important of the two problems, but takes longer to distinguish the probabilities of success and of failure due to the taxi moving, which are closer to each other.

## 6 DISCUSSION AND FUTURE DIRECTIONS

I have described a technique based on classical AI planning that is able to make use of probabilistic knowledge of external events to build more robust plans. It can search for a plan that passes a threshold for probability of success, or try to maximise this probability. AI planning brings a focus to the search based on goal-directed backward chaining that becomes increasingly important in more complex domains. To illustrate this point, consider how a planner based on policy iteration, such as used in [Dean *et al.*, 1993], would approach this problem. Policy iteration [Howard, 1960] computes the optimal action for each state, given the set of all states and a mapping from an action and a state to a probability distribution of successor states. In order to make policy iteration tractable, Dean et al. restrict attention to an envelope of the states contained in plans generated by an initial planner. Various schemes for incrementally increasing the envelope are considered, all based on adding a subset of the fringe, or the set of states that can be reached with some probability by executing the current plan. None of these schemes would be able to solve the worked example of this paper, since states with open lockers never appear on the fringe. In addition this path only shows greater utility when the two-step plan of opening the locker and storing the package is considered. The search space for considering all $n$-step plans from some envelope in a forward chaining manner would quickly become prohibitive, making goal-directed reasoning in some form necessary to generate envelopes for policy iteration in a moderately complex domain.

There are many interesting directions for this research. While the use of belief nets to criticise a plan shows promise, techniques to improve the tractability of the inference will be crucial to its success. This will be possible by exploiting the constraints imposed by the plan on the belief net as well as using techniques such as those described here to simplify its evaluation. In addition better use could be made of knowledge about dependencies between the events in the domain. The two approaches described in section 5 lie at opposite extremes, one requiring no information about independence and one assuming almost complete independence.

There are also benefits to be gained from relaxing the hard abstraction barrier between the planner and the routines that consider the effects of external events. Currently, the external events are viewed only as potential problems for the planner, rather than opportunities. The flexibility of the planner could be increased by incorporating these.

**Acknowledgements**

Jaime Carbonell provided many useful suggestions for this work, and discussions with Lonnie Chrisman and members of the Prodigy group were very helpful.

This research is sponsored by the Wright Laboratory, Aeronautical Systems Center, Air Force Materiel Command, USAF, and the Advanced Research Projects Agency (ARPA) under grant number F33615-93-1-1330. Views and conclusions contained in this document are those of the authors and should not be interpreted as necessarily representing official policies or endorsements, either expressed or implied, of Wright Laboratory or the United States Government.

This manuscript is submitted for publication with the understanding that the U.S. Government is authorized to reproduce and distribute reprints for Governmental purposes, notwithstanding any copyright notation thereon.